\documentclass[conference]{IEEEtran}
\usepackage{graphicx}
\usepackage[utf8]{inputenc} 
\usepackage{ulem}
\usepackage{amsmath}
\usepackage{epstopdf}
\usepackage{color}
\usepackage{hyperref}

\usepackage{bm}
\usepackage{amssymb}
\usepackage{extarrows}
\ifCLASSINFOpdf
\else
\fi
\begin{document}
%
\title{A Dual Sparse Decomposition Method\\ for Image Denoising}

\author{\IEEEauthorblockN{Hong Sun}
\IEEEauthorblockA{$^{1}$School of Electronic Information\\Wuhan University\\430072 Wuhan, China\\
$^{2}$Dept. Signal and Image Processing\\
Telecom ParisTech\\46 rue Barrault, 75013 Paris, France\\
Email: hongsun@whu.edu.cn}
\and
\IEEEauthorblockN{Chen-guang Liu}
\IEEEauthorblockA{$^{1}$School of Electronic Information\\Wuhan University\\430072 Wuhan, China\\
$^{2}$Dept. Signal and Image Processing\\
Telecom ParisTech\\46 rue Barrault, 75013 Paris, France\\
Email: chenguang.liu@telecomparistech.fr}
\and
\IEEEauthorblockN{Cheng-wei Sang}
\IEEEauthorblockA{School of Electronic Information\\Wuhan University\\430072 Wuhan, China\\
Email: sangcw@whu.edu.cn}}


%


\maketitle

\begin{abstract}
This article addresses the image denoising problem in the situations of strong noise. We propose a dual sparse decomposition method. This method makes a sub-dictionary decomposition on the over-complete dictionary in the sparse decomposition. The sub-dictionary decomposition makes use of a novel criterion based on the occurrence frequency of atoms of the over-complete dictionary over the data set. The experimental results demonstrate that the dual-sparse-decomposition method surpasses state-of-art denoising performance in terms of both peak-signal-to-noise ratio and structural-similarity-index-metric, and also at subjective visual quality.
\end{abstract}


%
\IEEEpeerreviewmaketitle

\section{Introduction}
Two main issues are involved in the denoising problem. One is the filtering technique by signal analysis to identify the information underlying the noisy data. The other is grouping technique by clustering technique to provide homogeneous signals for filtering.

Almost all filtering techniques assume that the involved signal should be homogeneous. Therefore, a grouping procedure is generally required before filtering. Many edge detection and image segmentation techniques \cite{IP_Henri} are used in image denoising. Recently, a nonlocal self-similarity method \cite{NonLocal} provides a potential breakthrough for data grouping, which is adopted in this paper.

The filtering technique is developed in the past 50 years or so from many diverse points of view, statistical estimation method, such as Viener filter, adaptive filter, etc. \cite{ADSP}; transform-domain method, such as Principal Components Analysis \cite{MathDSP}, wavelet shrinkage \cite{wavelet}, etc., and so on. The underlying assumption of these filtering methods is that information in the noisy data has a property of energy concentration in a small linear subspace of the overall space of possible data vectors, whereas additive noise is typically distributed through the larger space isotropically.

However, in many practical cases, some components with low energy might actually be important because they carry information relative to the signal details. On the contrary, when dealing with noise with non-Gaussian statistics, it may happen that some noise components may have higher energies. Consequently, a major difficulty of filtering is to separate the information details from noise. A way to deal with this problem is cooperative filtering technique, such as Turbo iterative filter \cite{Turbo}.

In recent years, sparse coding has attracted significant interest in the field of signal denoising \cite{Sparse} upon an over-complete dictionary. A sparse representation is a signal decomposition on a very small set of components (called atoms) which are adapted to the observational data. The sparse-decomposition based denoising is much better at the trade-off between the preservation of details and the suppression of noise. However, the sparse decomposition is adapted to noisy data so that separating details from noise still is at issue.

In this paper, we propose a dual sparse decomposition method for filtering. The first decomposition is to make an over-complete dictionary to reject some noises which really distributed through the larger space isotropically but to preserve the information details as much as possible. The second decomposition is to identify principal atoms to form a sub-dictionary which preserve well the weak information details and simultaneously suppress strong noises.

This article is organized as follows: Section 2 analyzes some limitations of the classical sparse decomposition for denoising. Section 3 presents the principle of the proposed dual sparse decomposition. Section 4 shows some experimental results and comparisons with state-of-art image denoising methods. Finally, we draw the conclusion in Section 5.

\section{Sparse Decomposition for Denoising}
We start with a brief description of the classical sparse decomposition and analyze their limitations for denoising.

The sparse decomposition of $M$ observations $\{\mathbf{x}_m \in
\mathbb{R}^N\}_{m=1}^M$ based on a dictionary
$\mathbf{D}=\{\mathbf{d}_k \}_{k=1}^K \in \mathbb{R}^{N \times K}$.
When $K>N$, the dictionary is said over-complete. $\mathbf{d}_k \in
\mathbb{R}^N$ is a basis vector, also called an atom of the dictionary. They are not
necessarily independent. With observational data set:
\begin{equation}\label{Eq_data}
\begin{aligned}
 \quad \mathbf{X}_{N \times M} = \{\mathbf{x}_m\}_{m=1}^M
\end{aligned}
\end{equation}
a dictionary and the coefficients can be the solution of the following equation \cite{KSVDalgo}:
\begin{equation}\label{Eq_1}
\begin{aligned}
  \{\mathbf{D},\bm{\alpha}_m\}=\operatorname*{arg min}\limits_{\mathbf{D},{\bm \alpha_m}} \parallel {\bm \alpha_m} \parallel_0 +  \parallel \mathbf{D}{\bm \alpha}_m - \mathbf{x}_m\parallel^2_2 \leq \varepsilon,\\ \quad 1 \leq m \leq M
   \end{aligned}
\end{equation}
where $\|\bullet\|_2$ denotes $\ell^2$-norm and $\|\bullet\|_0$ denotes $\ell^0$-norm. In equation (\ref{Eq_1}), $\bm{\alpha}_m=\left[ \alpha_m(1) \; \alpha_m(2) \; \dots \;
\alpha_m({K}) \right]^T \in \mathbb{R}^{K \times 1}$ is the sparse
code of the observation $\mathbf{x}_m$. The allowed error tolerance
$\varepsilon$ can be chosen according to the standard deviation of
the noise. The sparse decomposition can be written in matrix form as:
\begin{equation}\label{Eq_3}
\begin{aligned}
 \quad \mathbf{X}_{N \times M} \approxeq \mathbf{D}_{N \times K} \mathbf{A}_{K \times M}
\end{aligned}
\end{equation}
where the matrix $\mathbf{A}$  of size $K \times M$ is composed of
$M$ sparse column vectors $\bm{\alpha}_m$:
\begin{equation*}\label{coef_col}
\begin{aligned}
 \mathbf{A}_{K \times M}=\left[\bm{\alpha}_1 \cdots \bm{\alpha}_k \cdots \bm{\alpha}_K \right]
\end{aligned}
\end{equation*}

An estimate of the underlying signal
$\mathbf{S}$  embedded in the observed data set $\mathbf{X}$ would be:
\begin{equation}\label{Eq_2}
\begin{aligned}
 \quad \mathbf{S}
 =\left[ \mathbf{d}_1 \quad \mathbf{d}_2 \cdots \mathbf{d}_k \cdots \mathbf{d}_K \right]& .
   \left[\bm{\alpha}_1 \quad \bm{\alpha}_2 \cdots \bm{\alpha}_m \cdots \bm{\alpha}_M \right] &
\end{aligned}
\end{equation}

The over-complete dictionary $\mathbf{D}$ in sparse decomposition can effectively capture the information patterns and reject white Gaussian noise patterns. However, we note that the learning algorithm for dictionary $\mathbf{D}$  by equation (\ref{Eq_1}) would fall into a dilemma of preserving weak derails and suppressing noise. On one hand, in order to suppress noise, the allowed error tolerance $\varepsilon$ in equation (\ref{Eq_1}) should be small enough. As a result, certain weak details would be lost. On the other hand, in order to capture weak details, $\varepsilon$ cannot be too small. Otherwise some atoms would be so noisy that degrade the denoising performance. Fig. 1 shows an example to show this situation. Taking a noisy image degraded by white noise with standard deviation $\sigma=35$ (Fig. 1a), we make two different dictionaries $\textbf{D}_H$ and $\textbf{D}_L$ (Fig. 1b) by solving equation (\ref{Eq_1}) with $\varepsilon=40$ and $\varepsilon=35$ respectively. We got two different retrieved images $\textbf{S}_H$ and $\textbf{S}_L$ (Fig. 1b) respectively by equation (\ref{Eq_2}). Intuitively, the noise is well suppressed in $\textbf{S}_H$ but some information details are lost. On the contrary, more details are reserved in $\textbf{S}_L$ but it is rather noisy.

Considering the above limitation of sparse-decomposition-based denoising, our idea of dual sparse decomposition is to make a two-stop sparse decomposition: The first step is to make an over-complete dictionary by learning from the observational data with a lower allowed error tolerance $\varepsilon$ according to equation (\ref{Eq_1}). Thereby, the obtained dictionary $\textbf{D}_L$ can capture more information details although it contains some noisy atoms. The second step is to make a sub-dictionary decomposition on $\textbf{D}_L$ to reject some atoms too noisy.
\begin{figure*}[!t]
  \begin{center}
  \includegraphics[width=0.95 \linewidth]{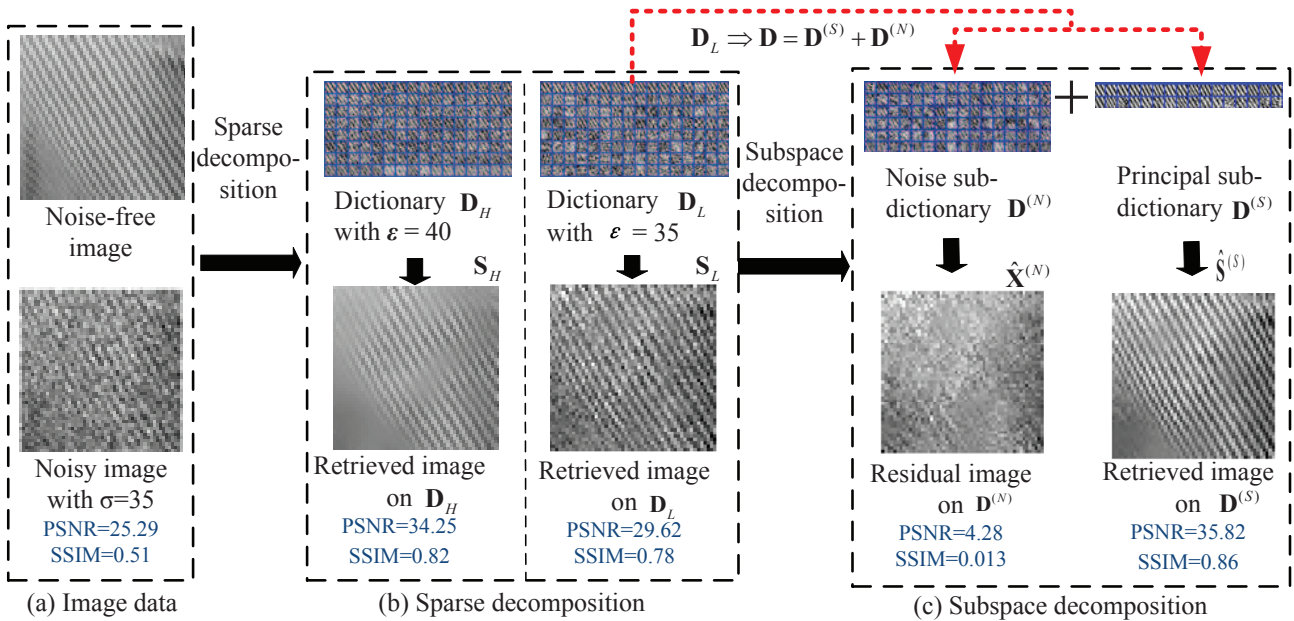}
 \caption{Principle of Dual Sparse decompositions.}\label{figure1}
  \end{center}
\end{figure*}

\section{Sparse Subspace Decomposition}
To get the sub-dictionary, we introduce a novel criterion to the sparse subspace decomposition of a learned dictionary and a corresponding index of significance of the atoms.

\subsection{Occurrence Frequency of Atom}
Atoms $\{\mathbf{d}_k \}_{k=1}^K$ in the sparse decomposition are prototypes of signal segments. This property allows us to take the atoms as a signal pattern. Thereupon, some important features of the signal pattern could be considered as a criterion to identify significant atoms. We note a common knowledge about the regularity of signal: A signal pattern must occur in meaningful signals with higher frequency even with a lower energy, such as the geometrical regularity of image structures like edges and textures. On the contrary, a noise pattern would hardly be reproduced in observed data even with a higher energy. Therefore, we propose to take the frequency of atoms appeared in the data set as the criterion to identify principal atoms \cite{PBA}. In fact, the frequency of atoms is a good description of the signal texture \cite{frequence}.

We intend to find out a measurement of the frequency of atom from the sparse codes. Coefficient matrix $\mathbf{A}$ in the sparse representation (equation (\ref{Eq_3})) is composed by $M$ sparse column vectors $\bm{\alpha}_m$. Let us consider the row vectors $\{\bm{\beta}_k\}_{k=1}^K$ of coefficient
matrix $\mathbf{A}$ :
\begin{equation*}\label{Eq_4}
   \begin{aligned}
 \mathbf{A}& = \left[\bm{\alpha}_1 \; \bm{\alpha}_2 \; \cdots \; \bm{\alpha}_M \right]\\& = \begin{bmatrix} \alpha_1(1) \quad \alpha_2(1) \quad \cdots \quad \alpha_M(1) \\ \alpha_1(2) \quad \alpha_2(2) \quad \cdots \quad \alpha_M(2)\\\vdots \quad \quad \quad \vdots \quad \quad \ddots \quad \quad \vdots \\ \alpha_1(K) \quad \alpha_2(K) \quad \cdots \quad \alpha_M(K)   \end{bmatrix} =  \begin{bmatrix} \bm{\beta}_1 \\ \bm{\beta}_2 \\ \vdots \\ \bm{\beta}_K \end{bmatrix}
   \end{aligned}
\end{equation*}
where
\begin{equation}\label{Eq_Vcol}
   \begin{aligned}
 \bm{\beta}_k = [\alpha_1(k)  \; \alpha_2(k) \; \dots \; \alpha_M(k)] \in \mathbb{R}^{1 \times M}
   \end{aligned}
\end{equation}
Note that the row vector $\bm{\beta}_k$ is not necessarily sparse.

Thus, the coefficient matrix $\mathbf{A}$ can be written by $K$ row vectors as:
\begin{equation*}\label{Eq_Crow}
\begin{aligned}
 \mathbf{A}_{K \times M}=\left[\bm{\beta}_1^T \cdots \bm{\beta}_k^T \cdots \bm{\beta}_K^T \right]^T
\end{aligned}
\end{equation*}

Then equation (\ref{Eq_2}) can be expressed by reordered dictionary and its coefficient as:
\begin{equation}\label{Eq_ResB}
 \begin{aligned}
   \mathbf{S}_{N \times M} = \left[\mathbf{d}_1 \cdots \mathbf{d}_k \cdots \mathbf{d}_K \right] . \left[\bm{\beta}_1^T \cdots \bm{\beta}_k^T \cdots \bm{\beta}_K^T \right]^T
 \end{aligned}
\end{equation}

Denoting $\|\bm{\beta}_k\|_0$ the $\ell^0$ zero pseudo-norm of $\bm{\beta}_k$, we find that $\|\bm{\beta}_k\|_0$ is just the number of occurrences of atom $\mathbf{d}_k$ over the data set
$\{\mathbf{x}_m \}_{m=1}^M$. We can define the frequency of the atom
$\mathbf{d}_k$ as $f_k$:
\begin{equation}\label{Eq_6}
 \begin{aligned}
   f_k\triangleq Frequency(\mathbf{d}_k | \mathbf{X}) = \| \bm{\beta}_k \|_0
 \end{aligned}
\end{equation}

\subsection{Subspace decomposition on Over-complete Dictionary}
Taking vectors $\{ \bm{\beta}_k\}_{k=1}^K$ from equation (\ref{Eq_Vcol}), we calculate their $\ell^0$-norms $\{\|\bm{\beta}_k \|_0\}_{k=1}^K$ and rank them in descending order:
\begin{equation}\label{Eq_7}
 \begin{aligned}
 \quad \tilde{\mathbf{A}} \triangleq & [\bm\beta'_1,\cdots,\bm\beta'_k,\cdots,\bm\beta'_K ] \xLongleftarrow{sort}  [\bm\beta_1,\cdots,\bm\beta_k,\cdots,\bm\beta_K ] \\ & \textit{s.t.} \parallel\bm\beta'_1\parallel_0 \geq \parallel\bm\beta'_2\parallel_0 \geq \cdots \geq \parallel\bm\beta'_K\parallel_0
\end{aligned}
\end{equation}

Corresponding to the order of $\{\bm \beta'_k \}_{k=1}^K$, the reordered dictionary is written as:
\begin{equation}\label{Eq_8}
 \begin{aligned}
 \mathbf{D} \xLongrightarrow{sort} \tilde{\mathbf{D}} =  [\textbf{d}'_1,\cdots,\textbf{d}'_k,\cdots,\textbf{d}'_K ]
\end{aligned}
\end{equation}
Equation (\ref{Eq_ResB}) becomes as:
\begin{equation}\label{Eq_sort}
 \begin{aligned}
   \mathbf{S}_{N \times M}& = \mathbf{D}_{N \times K} \mathbf{A}_{K \times M} = \tilde{\mathbf{D}}_{N \times K} \tilde{\mathbf{A}}_{K \times M} \\& = \left[\mathbf{d}'_1, \cdots,\mathbf{d}'_k, \cdots,\mathbf{d}'_K \right] . \left[ \bm{\beta}_1^{'T} \cdots \bm{\beta}_k^{'T} \cdots \bm{\beta}_K^{'T} \right]^T
 \end{aligned}
\end{equation}

Then, the first $P$ atoms of $\tilde{\mathbf{D}}$ can span a principal
subspace $\mathbf{D}_P^{(S)}$ and the remaining atoms span
a noise subspace $\mathbf{D}_{K-P}^{(N)}$ as:
\begin{equation}\label{Eq_sub}
\begin{aligned}
  \mathbf{D}_P^{(S)} = span\{\mathbf{d}'_1,\mathbf{d}'_2,\cdots,\mathbf{d}'_P \} \qquad   \\ \mathbf{D}_{K-P}^{(N)} = span\{\mathbf{d}'_{P+1},\mathbf{d}'_{P+2},\cdots,\mathbf{d}'_K \}
   \end{aligned}
\end{equation}

In practical application, $P$ is the threshold of $f_k$ to separate the principal sub-dictionary from the noise sub-dictionary. We set the maximum point of the histogram of $\{\|\bm{\beta}_k \|_0\}_{k=1}^K$ to $P$ as:
\begin{equation}\label{Eq_cut}
\begin{aligned}
 P =\operatorname*{arg \ max \ Hist}\limits_{k}{(\|\bm\beta'_k \|_0)}
 \end{aligned}
\end{equation}

An estimate of the underlying signal $\hat{\mathbf{S}}$ embedded
in the observed data set $\mathbf{X}$ can be obtained on the
principal sub-dictionary $\mathbf{D}^{(\mathbf{S})}$ simply by linear
combination:
\begin{equation}\label{Eq_PC}
\begin{aligned}
 \hat{\mathbf{S}} &= \mathbf{D}_P^{(S)} . \mathbf{A}_P^{(S)}  \\
  &=  \left[\mathbf{d}'_1, \cdots,\mathbf{d}'_k, \cdots,\mathbf{d}'_P  \right] . \left[\bm{\beta}_1^{'T} \cdots \bm{\beta}_k^{'T} \cdots \bm{\beta}_P^{'T} \right]^T
\end{aligned}
\end{equation}
Note that $P\ll K$.

We show an example of the proposed dual sparse decomposition in Fig. 1(c). The learned over-complete dictionary $\mathbf{D}$ is decomposed into a principal sub-dictionary $\mathbf{D}^{(S)}$ and a noise sub-dictionary $\mathbf{D}^{(N)}$ under the atom's frequency criterion. The retrieved image $\hat{\mathbf{S}}^{(S)}$ by the dual sparse decomposition method has a super performance at preserving fine details and at suppressing strong noise. We note that the residual image $\mathbf{X}^{(N)}$ on the noise sub-dictionary $\mathbf{D}^{(N)}$ contains some information but very noisy. This is because the atoms of the over-complete dictionary are not independent. The information in the residue image $\mathbf{X}^{(N)}$ is also in existence in $\hat{\mathbf{S}}^{(S)}$.

\subsection{Application to Filtering}
A major difficulty of filtering is to suppress noise Gaussian or non-Gaussian and to preserve information details simultaneously. We use the peak signal-to-noise ratio ($\textit{PSNR}$) to assess the noise removal performance:
\begin{equation*}\label{Eq_13}
 \begin{aligned}
\textit{P}&\textit{SNR}=20 \cdot \log_{10} \left[\max\{\mathbf{S}(i,j) \}\right] -10
  \cdot \log_{10} \left[ \textit{MSE} \right] \\
&\textit{MSE}=\frac{1}{IJ}{\sum\nolimits_{i=0}^{I-1}}{\sum\nolimits_{j=0}^{J-1}}\left[\mathbf{S}(i,j)-\hat{\mathbf{S}}(i,j)
  \right]^2
 \end{aligned}
\end{equation*}
and the structural similarity index metric ($\textit{SSIM}$) between denoised image $\hat{\mathbf{S}}$ and the pure one $\mathbf{S}$ to evaluate the preserving details performance:
\begin{equation*}\label{Eq_14}
 \begin{aligned}
\textit{SSIM}(\mathbf{S},\hat{\mathbf{S}})=\frac{(2u_{\mathbf{S}}u_{\hat{{\mathbf{S}}}} + c_1)(2\sigma_{\mathbf{S}\hat{{\mathbf{S}}}}+c_2) }{(u^2_{\mathbf{S}}+ u^2_{\hat{\mathbf{S}}} +c_1)(\sigma^2_{\mathbf{S}} + \sigma^2_{\hat{\mathbf{S}}}+ c_2) }
 \end{aligned}
\end{equation*}
where $u_x$ is the average of $x$, $\sigma_x^2$ is the variance of $x$, $\sigma_{xy}$ is the covariance of $x$ and $y$, and $c_1$ and $c_2$ are small variables to stabilize the division with weak denominator.

From the example shown in Fig. 1, the retrieved image $\mathbf{S}_H$ actually by the K-SVD filter \cite{KSVDfilter} with the classical sparse decomposition has a high performance with $\textit{PSNR}=34.25$ and $\textit{SSIM}=0.82$ but some information details are obviously lost.  On the contrary, the retrieved image $\mathbf{S}_L$ is noisier with $\textit{PSNR}=29.62$ and $\textit{SSIM}=0.78$ but more information details are reserved. Making a dictionary decomposition on $\mathbf{D}_L$ noisier but with more details, the retrieved image $\hat{\mathbf{S}}^{(S)}$ based on the principal sub-dictionary $\mathbf{D}^{(S)}$ has a higher performance with $\textit{PSNR}=35.82$ and $\textit{SSIM}=0.86$.

Fig. 2 shows an image filtering result based on the proposed dual sparse decomposition and a comparison with K-SVD algorithm. From the results, the dual sparse decomposition method outperforms K-SVD method by about $1dB$ in $\textit{PSNR}$ and by about $1\%$ in $\textit{SSIM}$. In terms of subjective visual quality, we can see that the corner of mouth and the nasolabial fold with weak intensities are much better recovered by the dual sparse decomposition method.
\begin{figure}[h!]
  \begin{center}
   \includegraphics[width=0.95 \linewidth]{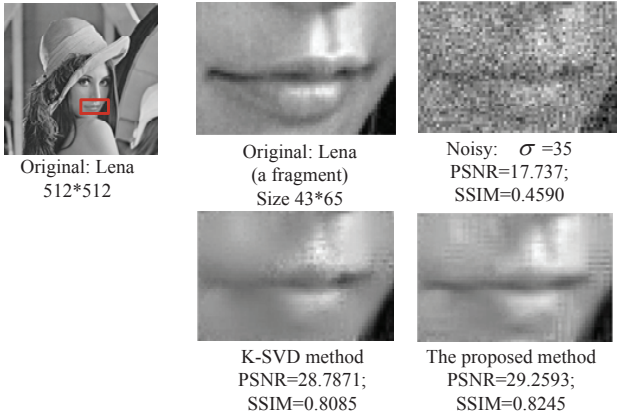}
   \caption{Image filtering by the dual sparse decomposition comparing with the K-SVD method.}\label{figure2}
  \end{center}
\end{figure}

Fig. 3 shows the despeckling results of simulated one-look SAR scenario with a fragment of Barbara image. From the result by a probabilistic patch based (PPB) filter \cite{PPB} which can cope with non-Gaussian noise, we can see that PPB can well remove speckle noise. However, it also removes low-intensity details. The dual sparse decomposition method shows advantages at preserving fine details and at suppressing strong noise.
\begin{figure}[h!]
  \begin{center}
   \includegraphics[width=0.95 \linewidth]{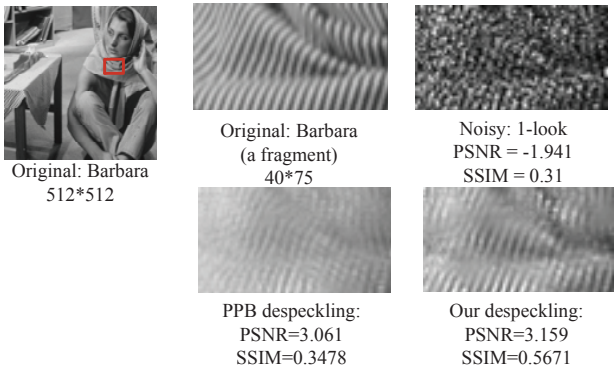}
   \caption{SAR image despeckling coparing the proposed dual sparse decomposition method with the PPB methode.}\label{figure3}
  \end{center}
\end{figure}

\section{Application to Denoising}

In practical applications, our images are generally with spatial complicated scene. On the other hand, the used filtering techniques are generally suitable to homogeneous images. For image denoising based on the sparse decomposition, the hypotheses of signal sparsity and component reproducibility mean also the condition of homogeneity. In order to make the involved signal homogeneous, we select homogeneous pixels before filtering by a self-similarity measure \cite{NonLocal} $\gamma$. In applications of image denoising, $\gamma$ can be specified as Euclidean distance between the reference patch $\mathbf{x}_i$ and a given patch $\mathbf{x}_j$ as:
\begin{equation}\label{Eq_15}
 \begin{aligned}
   \gamma(\mathbf{x}_i,\mathbf{x}_j)  = \| \mathbf{x}_i - \mathbf{x}_j \|_0^{2}
 \end{aligned}
\end{equation}
The smaller $\gamma$ is, the more similar between $\mathbf{x}_i$ and $\mathbf{x}_j$ is. This self-similarity matches well the property of highly repetitive structures of images.

In applications of image despeckling, $\gamma$ becomes the probabilistic patch-based similarity proposed by \cite{PPB} as:
\begin{equation}\label{Eq_16}
 \begin{aligned}
   \gamma(\mathbf{x}_i,\mathbf{x}_j)  = (2L-1)\sum_{k}{\log{\sqrt{\frac{\mathbf{y}_i(k)}{\mathbf{y}_j(k)}}+\sqrt{\frac{\mathbf{y}_j(k)}{\mathbf{y}_i(k)}}}}
 \end{aligned}
\end{equation}
where $\mathbf{y}_i=\exp(\mathbf{x}_i)$ and $L$ the equivalent number of looks.

\begin{figure}[h!]
  \begin{center}
   \includegraphics[width=0.95 \linewidth]{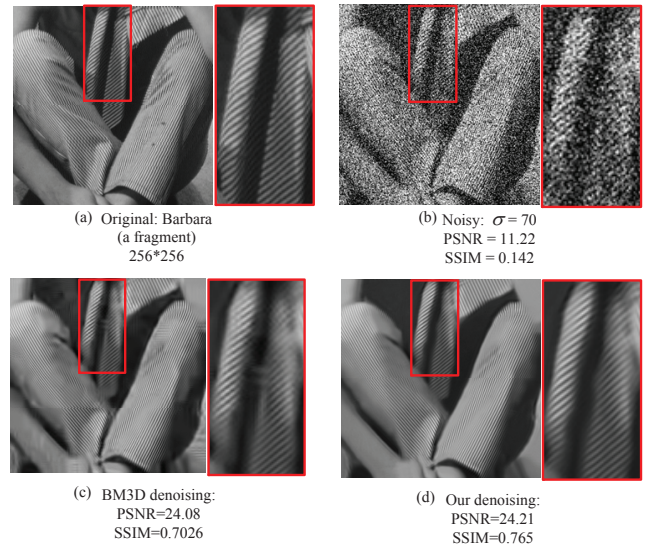}
   \caption{Denoising for spatial complicated image scene comparing BM3D method.}\label{figure4}
  \end{center}
\end{figure}

\begin{figure*}[!t]
  \begin{center}
  \includegraphics[width=0.95 \linewidth]{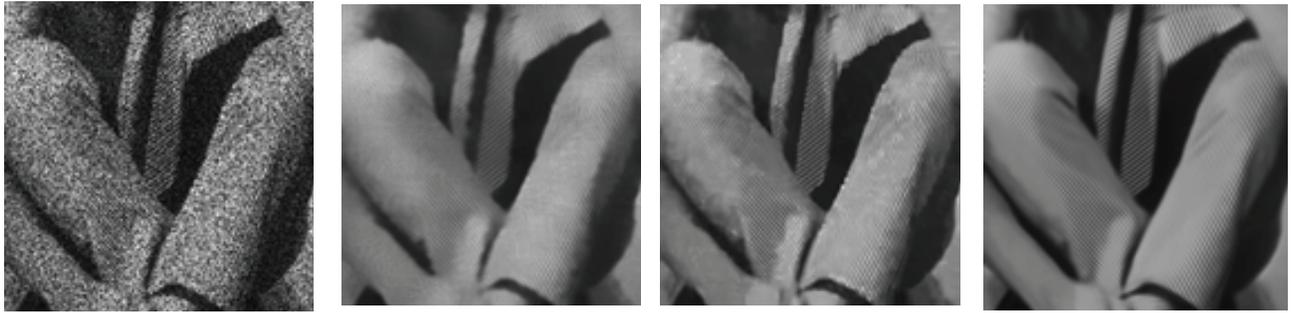}
 \caption{Principle of Dual Sparse decompositions.}\label{figure5}
  \end{center}
\end{figure*}

For a given reference patch $\mathbf{x}_i$, we make grouping stacks with its $\Gamma$-most similar patches to form a group of data $\mathbf{z}_i$. In our experiments, we take $\Gamma=90$. Then we apply a filtering algorithm to each of the data groups $\mathbf{z}_i,\forall i$. Our denoising algorithm is presented in Table \ref{Tab_Alg}:

\begin{table}
 \centering
 \caption{Denoising Algorithm Based on Dual Sparse Decomposition}\label{Tab_Alg}
\begin{tabular*}{8.8cm}{@{}l @{} l }
  \hline
  Input: Image data $\mathbf{X}= \{\mathbf{x}_m
\}_{m=1}^M$ (Equ.(\ref{Eq_data})\\
  Grouping: For patch $\mathbf{x}_{i}$, form group $\mathbf{z}_{i}$ according Equs. (\ref{Eq_15}) or (\ref{Eq_16})\\
  Dual sparse decomposition: For each group $\mathbf{z}_{i},\forall i$ do\\
  - Sparse decomposition: $\mathbf{z}_{i} \approxeq \mathbf{D} \mathbf{A}$ by solving Equ.(\ref{Eq_1})\\
  - Subspace decomposition: $\mathbf{D}=\mathbf{D}^{(\mathbf{S})}+\mathbf{D}^{(\mathbf{N})}$ by Equs.(\ref{Eq_Vcol}, (\ref{Eq_7})-(\ref{Eq_sub})\\
  - Linear reconstruction on $\mathbf{D}^{(\mathbf{S})}$: $\hat{\mathbf{S}}_{i} = \mathbf{D}_P^{(\mathbf{S})} . \mathbf{A}_P^{(\mathbf{S})}$ by Equs. (\ref{Eq_cut})-(\ref{Eq_PC})\\
  Aggregate: to form denoised image  $weight-average\{\hat{\mathbf{S}}_{i}, \forall i \} \Rightarrow \hat{\mathbf{S}}$\\
  Output: Denoised image $\hat{\mathbf{S}}$.\\
  \hline
\end{tabular*}
\end{table}

To compare with the state-of-art denoising algorithm, we take the BM3D method \cite{BM3D}, one of the best method nowadays for image denoising. In the BM3D method, a block-matching grouping is also used before filtering. In the experiments, the used dictionaries $\mathbf{D}$s are of size are of size $64 \times 256$ ($K=256$ atoms), designed to handle image patches $\mathbf{x}_m$ of size $N=64=8 \times 8$ pixels.

Fig. 4 shows the results of denoising an image with a strong additive zero-mean white Gaussian noise and their performances of the dual-sparse-decomposition method and the BM3D method. Fig. 5 shows the results of despeckling a simulated one-look SAR image with non-Gaussian noise and their performances of the dual-sparse-decomposition method and the SAR-BM3D method \cite{SAR}. The experimental results demonstrate some advantage of the dual-sparse-decomposition method at preserving fine details and at suppressing speckle noise, also with a better subjective visual quality over the BM3D method.

\section{Conclusion}
This work present a new signal analysis method by a proposed dual sparse decomposition, leading to state-of-the-art performance for image denoising. The proposed method introduces a sub-dictionary decomposition on an over-complete dictionary learned under a lower allowed-error-tolerance. The principal sub-dictionary is identified under a novel criterion based on the occurrence frequency of atoms. The experimental results have demonstrated that the proposed dual-sparse-decomposition-based denoising method has some advantages both at preserving information details and at suppressing strong noise, as well as provides retrieved image with better subjective visual quality.

It is perfectly possible to straightforward extension the proposed dual-sparse-decomposition to application of feature extraction, inverse problems, or machine learning.


\section*{Acknowledgment}
This work was supported by the National Natural Science Foundation of China (Grant No. 60872131).

The idea of the dual Sparse decomposition arises through a lot of deep discussions with Professor Henri Ma$\hat{i}$tre at Telecom-ParisTech. The mathematical expressions in this paper are corrected by Professor Didier Le Ruyet at CNAM-Paris.






%

%

\bibliographystyle{IEEEtran}
\bibliography{MyBibfile}



\end{document}